\documentclass[conference]{IEEEtran}
\IEEEoverridecommandlockouts
\usepackage{cite}
\usepackage{amsmath,amssymb,amsfonts}
\usepackage{graphicx}
\usepackage{textcomp}
\usepackage{multirow}
\usepackage{booktabs}
\usepackage{comment}
\usepackage{tikz}
\usepackage{pgfplots}
\usepackage{caption}
\usepackage{color, colortbl}
\usepackage{subcaption}
\usepackage[dvipsnames]{xcolor}
\usepackage{url}

\usepackage{comment}

\usepackage{nohyperref}

\definecolor{matplotlib0}{HTML}{1f77b4}
\definecolor{matplotlib1}{HTML}{d62728}
\definecolor{matplotlib2}{HTML}{2ca02c}
\definecolor{matplotlib3}{HTML}{ff7f0e}
\definecolor{matplotlib4}{HTML}{9467bd}
\definecolor{matplotlib5}{HTML}{8c564b}
\definecolor{matplotlib6}{HTML}{e377c2}
\definecolor{matplotlib7}{HTML}{7f7f7f}
\definecolor{matplotlib8}{HTML}{bcbd22}
\definecolor{matplotlib9}{HTML}{17becf}

\usepackage{mathtools} 

\usepackage{subcaption}

\usepackage{booktabs}
\usepackage{array}
\usepackage{multirow}
\usepackage{colortbl}
\usepackage{tablefootnote}
\usepackage{threeparttable}

\usepackage[acronym, style=super, nonumberlist]{glossaries}

\usepackage{pgfplots}
\definecolor{color0}{rgb}{0.12156862745098,0.466666666666667,0.705882352941177} 
\definecolor{color1}{rgb}{1,0.498039215686275,0.0549019607843137}
\definecolor{color2}{rgb}{0.172549019607843,0.627450980392157,0.172549019607843} 
\definecolor{color3}{rgb}{0.83921568627451,0.152941176470588,0.156862745098039} 
\definecolor{color4}{rgb}{0.580392156862745,0.403921568627451,0.741176470588235}
\definecolor{colorblue}{rgb}{0.12156862745098,0.466666666666667,0.705882352941177} 
\definecolor{colorgreen}{rgb}{0.172549019607843,0.627450980392157,0.172549019607843} 
\definecolor{colorred}{rgb}{0.83921568627451,0.152941176470588,0.156862745098039} 
\definecolor{colorblack}{rgb}{0,0,0} 
\definecolor{colororange}{rgb}{1,0.56,0} 
\usepgfplotslibrary{fillbetween}
\usepgfplotslibrary{colormaps}
\pgfplotsset{compat=1.16}

\pgfplotscreateplotcyclelist{matplotlib}{
  {matplotlib0},
  {matplotlib1},
  {matplotlib2},
  {matplotlib3},
  {matplotlib4},
  {matplotlib5},
  {matplotlib6},
  {matplotlib7},
  {matplotlib8},
  {matplotlib9}
}

\pgfplotsset{every axis/.append style={
    cycle list name=matplotlib,
}}

\usepackage{listings}

\definecolor{code_default}{HTML}{000000}
\definecolor{code_keyword}{HTML}{AC4142}
\definecolor{code_identifier}{HTML}{D28445}

\lstdefinelanguage{RISCV}{
  sensitive=false,
  morecomment=[l]{//},
  alsoletter={.},
  morekeywords=[1]{
    lp.setup, mv, lw, p.lw, sw, p.sw, pv.sdotsp.b, pv.shuffle2.b, p.subNR, p.addNR
  },
  morekeywords=[2]{
    zero, ra, sp, gp, tp, t0, t1, t2, t3, t4, t5, t6, s0, s1, a0, a1, a2, a3, a4, a5, a6, a7, a8, a9, a10, a11,
  },
  morestring=[b]",
  morestring=[b]',
}[strings, comments, keywords]

\lstdefinestyle{RISCV_STYLE}{
  language=RISCV,
  numbers=none,
  basicstyle=\scriptsize\ttfamily\color{code_default},
  keywordstyle=[1]\color{matplotlib0},
  keywordstyle=[2]\color{matplotlib1},
  float,
  captionpos=b,
  belowskip=-0.5cm
}

\usepackage{algorithm}
\usepackage{algpseudocode}
\usepackage{float}
\newfloat{algorithm}{t}{top}

\newacronym{simd}{SIMD}{Single Instruction, Multiple Data}
\newacronym{elu}{ELU}{Exponential Linear Unit}
\newacronym{relu}{ReLU}{Rectified Linear Unit}
\newacronym{gelu}{GELU}{Gaussian Error Linear Unit}
\newacronym{rpr}{RPR}{Random Partition Relaxation}
\newacronym{mac}{MAC}{Multiply Accumulate}
\newacronym{dma}{DMA}{Direct Memory Access}
\newacronym{bmi}{BMI}{Brain--Machine Interface}
\newacronym{bci}{BCI}{Brain--Computer Interface}
\newacronym{smr}{SMR}{Sensory Motor Rythms}
\newacronym{eeg}{EEG}{Electroencephalography}
\newacronym{svm}{SVM}{Support Vector Machine}
\newacronym{svd}{SVD}{Singular Value Decomposition}
\newacronym{evd}{EVD}{Eigendecomposition}
\newacronym{iir}{IIR}{Infinite Impulse Response}
\newacronym{fir}{FIR}{Finite Impulse Response}
\newacronym{fc}{FC}{Fabric Controller}
\newacronym{nn}{NN}{Neural Network}
\newacronym{mrc}{MRC}{Multiscale Riemannian Classifier}
\newacronym{flop}{FLOP}{Floating Point Operation}
\newacronym{sos}{SOS}{Second-Order Section}
\newacronym{ipc}{IPC}{Instructions per Cycle}
\newacronym{tcdm}{TCDM}{Tightly Coupled Data Memory}
\newacronym{fpu}{FPU}{Floating Point Unit}
\newacronym{fma}{FMA}{Fused Multiply Add}
\newacronym{alu}{ALU}{Arithmetic Logic Unit}
\newacronym{dsp}{DSP}{Digital Signal Processing}
\newacronym{gpu}{GPU}{Graphics Processing Unit}
\newacronym{soc}{SoC}{System-on-Chip}
\newacronym{mi}{MI}{Motor-Imagery}
\newacronym{csp}{CSP}{Commmon Spatial Patterns}
\newacronym{fbcsp}{FBCSP}{Filter-Bank \acrlong{csp}}
\newacronym{pulp}{PULP}{parallel ultra-low power}
\newacronym{soa}{SoA}{state-of-the-art}
\newacronym{bn}{BN}{Batch Normalization}
\newacronym{isa}{ISA}{Instruction Set Architecture}
\newacronym{ecg}{ECG}{Electrocardiogram}
\newacronym{mcu}{MCU}{microcontroller}
\newacronym{rnn}{RNN}{recurrent neural network}
\newacronym{cnn}{CNN}{convolutional neural network}
\newacronym{tcn}{TCN}{temporal convolutional network}
\newacronym{emu}{EMU}{epilepsy monitoring unit}
\newacronym{pd}{PD}{Parkinson's disease}
\newacronym{ad}{AD}{Alzheimer's disease}
\newacronym{tueg}{TUEG}{Temple University Hospital EEG}
\newacronym{tuh}{TUH}{Temple University Hospital}
\newacronym{tusl}{TUSL}{TUH EEG Slowing Corpus}
\newacronym{tuar}{TUAR}{TUH EEG Artifact Corpus}
\newacronym{tuab}{TUAB}{TUH Abnormal EEG Corpus}

\newacronym{fm}{FM}{Foundation Model}
\newacronym{nlp}{NLP}{natural language processing}
\newacronym{gpt}{GPT}{Generative Pre-trained Transformer}
\newacronym{ssm}{SSM}{State Space Model}
\newacronym{moe}{MoE}{Mixture of Experts}
\newacronym{bimamba}{bi-Mamba}{bidirectional Mamba}
\newacronym{lejepa}{LeJEPA}{Latent-Euclidean Joint-Embedding Predictive Architecture}
\newacronym{jepa}{JEPA}{Joint-Embedding Predictive Architecture}
\newacronym{sigreg}{SIGReg}{Sketched Isotropic Gaussian Regularization}
\newacronym{ecf}{ECF}{Empirical Characteristic Function}
\newacronym{cf}{CF}{Characteristic Function}
\newacronym{tsne}{t-SNE}{T-distributed Stochastic Neighbor Embedding}
\newacronym{flops}{FLOPS}{Floating-point Operations Per Second}
\newacronym{oom}{OOM}{Out-Of-Memory}
\newacronym{ssl}{SSL}{Self-Supervised Learning}
\newacronym{fft}{FFT}{Fast Fourier Transform}

\newacronym{lda}{LDA}{Linear Discriminant Analysis}
\newacronym{sota}{SOTA}{state-of-the-art}
\newacronym{auroc}{AUROC}{Area Under the Receiver Operating Characteristic}
\newacronym{aupr}{AUPR}{Area Under the Precision-Recall}
\newacronym{balacc}{Bal. Acc}{Balanced Accuracy}

\renewcommand{\baselinestretch}{0.95}
\def\BibTeX{{\rm B\kern-.05em{\sc i\kern-.025em b}\kern-.08em
    T\kern-.1667em\lower.7ex\hbox{E}\kern-.125emX}}
\begin{document}
\thispagestyle{empty}
\pagestyle{empty}

\title{
LuMamba: Latent Unified Mamba for Electrode Topology-Invariant and Efficient EEG Modeling}

 \author{\IEEEauthorblockN{
    Danaé Broustail\IEEEauthorrefmark{1},
    Anna Tegon\IEEEauthorrefmark{1},
    Thorir Mar Ingolfsson\IEEEauthorrefmark{1},
    Yawei Li\IEEEauthorrefmark{1}\IEEEauthorrefmark{2},
    Luca Benini\IEEEauthorrefmark{1}\IEEEauthorrefmark{3}}
    \IEEEauthorblockA{\IEEEauthorrefmark{1}Integrated Systems Laboratory, ETH Z{\"u}rich, Z{\"u}rich, Switzerland}
    \IEEEauthorblockA{\IEEEauthorrefmark{2}School of Electrical and Electronic Engineering, Nanyang Technological University, Singapore}
    \IEEEauthorblockA{\IEEEauthorrefmark{3}DEI, University of Bologna, Bologna, Italy}
    \thanks{Corresponding email: \{ategon,thoriri\}@iis.ee.ethz.ch}
    }
\maketitle

\begin{abstract}
Electroencephalography (EEG) enables non-invasive monitoring of brain activity across clinical and neurotechnology applications, yet building foundation models for EEG remains challenging due to differing electrode topologies and computational scalability, as Transformer architectures incur quadratic sequence complexity. As a joint solution, we propose LuMamba (Latent Unified Mamba), a self-supervised framework combining topology-invariant encodings with linear-complexity state-space modeling, using LUNA's learned-query cross-attention mechanism for channel unification, and FEMBA's bidirectional Mamba blocks for efficient temporal modeling. Within this architecture, we provide the first systematic investigation of the Latent-Euclidean Joint-Embedding Predictive Architecture (LeJEPA) for biosignal learning. Pre-trained on over 21,000 hours of unlabeled EEG from the TUEG corpus, LuMamba is evaluated on five downstream tasks spanning abnormality detection, artifact recognition, and mental condition classification across electrode configurations ranging from 16 to 26 channels. In the pre-training objective, masked reconstruction alone yields structured but less generalizable representations, while LeJEPA alone produces diffuse embeddings; combining both objectives achieves the most robust performance. With only 4.6M parameters, LuMamba attains 80.99\% balanced accuracy on TUAB and achieves state-of-art performance on Alzheimer's detection (0.97 AUPR), while requiring 377$\times$ fewer FLOPS than state-of-art models at equivalent sequence lengths and scaling to 12$\times$ longer sequences before reaching typical GPU memory limits. Code is available at \url{https://github.com/pulp-bio/biofoundation}.

\end{abstract}

\section{Introduction}\label{sec:intro}

\gls{eeg} provides non-invasive access to brain activity and plays a central role in clinical diagnostics, cognitive neuroscience, and brain--computer interfaces. The emergence of foundation models has transformed \gls{eeg} analysis, enabling \gls{ssl} on large unlabeled corpora followed by task-specific fine-tuning~\cite{bendr,labram,biot}. While Transformer-based approaches have dominated this space~\cite{cerebro,eegformer_foundation}, their quadratic complexity in sequence length poses challenges for long \gls{eeg} recordings and resource-constrained deployment \cite{luna, femba}. \glspl{ssm}, particularly Mamba~\cite{mamba}, offer a compelling alternative: FEMBA~\cite{femba} demonstrates that bidirectional \glspl{ssm} can match Transformer performance on clinical benchmarks while reducing computational cost by up to $3.5\times$.

However, a fundamental bottleneck persists: \emph{topological heterogeneity}. Electrode count and placement vary widely across \gls{eeg} datasets, causing pronounced performance degradation during cross-montage transfer. That is, models can lose 2-6\% when evaluated on unseen electrode configurations~\cite{luna}, while approaches that retain only shared electrodes discard substantial portions of available data~\cite{cerebro}. LUNA~\cite{luna} addresses this via learned-query cross-attention, projecting electrode layouts into a fixed latent space, albeit at the cost of a computationally intensive Transformer encoder. Whether such topology-invariant encoding remains effective when combined with efficient \gls{ssm} backbones is an open question.

A further gap lies in the pre-training objective itself. Current \gls{eeg} foundation models predominantly rely on masked reconstruction~\cite{femba,luna,labram} or contrastive learning~\cite{bendr}, yet the optimal strategy for \gls{ssm} architectures remains unclear. Recently, \gls{lejepa}~\cite{balestriero2025lejepaprovablescalableselfsupervised} emerged as a theoretically grounded alternative that regularizes embeddings toward an isotropic Gaussian distribution while aligning local and global views, but it has not yet been applied to biosignal time series.

To address these gaps, we introduce \textbf{LuMamba} (\textbf{L}atent \textbf{U}nified \textbf{Mamba}), a self-supervised framework that unifies topology-invariant encoding with linear-complexity state-space modeling. LuMamba fuses LUNA's channel-unifying cross-attention with FEMBA's efficient \gls{bimamba} blocks, enabling---for the first time---evaluation of \gls{ssm} architectures across heterogeneous electrode configurations. Within this framework, we systematically investigate the interplay between \gls{lejepa} and masked reconstruction. Our experiments reveal that while \gls{lejepa} alone produces embeddings with poor clustering structure, and reconstruction alone yields well-structured embeddings with lower cross-montage robustness, combining both objectives achieves the best of both worlds: strong downstream performance \emph{and} improved generalization to unseen electrode configurations.
Our contributions are the following:
\begin{itemize}
    \item \textbf{Topology-Invariant \gls{ssm} for \gls{eeg}:}
    We introduce \textit{LuMamba}, a fused architecture combining FEMBA's \gls{bimamba} backbone with LUNA's channel-unifying cross-attention, enabling topology-invariant \gls{eeg} modeling across heterogeneous montages (16--26 channels).

    \item \textbf{First Adaptation of \gls{lejepa} to Biosignals:}
    We adapt the \gls{lejepa} framework to \gls{eeg} time series via temporal global/local view extraction, and characterize its interaction with masked reconstruction pre-training.

    \item \textbf{SSL Objective Trade-offs:}
    We compare reconstruction-only, LeJEPA-only, and mixed pre-training objectives, revealing a trade-off between latent structure compactness and downstream generalization, especially under distribution shifts and unseen electrode configurations. On \gls{ad} detection, the mixed objective improves \gls{aupr} by over \textbf{20\%} compared to reconstruction alone.

    \item \textbf{Efficiency and Scalability:}
    With only 4.6M parameters, LuMamba requires \textbf{26$\times$ fewer \gls{flops}} than LUNA and \textbf{377$\times$ fewer} than LaBraM at equivalent sequence lengths, scaling to \textbf{12$\times$ longer sequences} before reaching memory limits.
\end{itemize}
\section{Background}\label{sec:related}
\paragraph*{Foundation Models for \gls{eeg}}
Foundation models have achieved remarkable success in \gls{nlp} and computer vision~\cite{bommasani2022opportunitiesrisksfoundationmodels}, motivating their application to \gls{eeg} analysis. Early approaches such as BENDR~\cite{bendr} introduced contrastive \gls{ssl} for large-scale \gls{eeg} data, while subsequent works refined masked modeling strategies. In particular, LaBraM~\cite{labram} and EEGFormer~\cite{eegformer_foundation} learn representations by predicting masked patches, and BIOT~\cite{biot} extends these ideas to cross-dataset learning. Such \gls{eeg} foundation models rely on Transformer-based architectures, motivating the exploration of more computationally efficient alternatives such as \glspl{ssm}~\cite{femba}.
\paragraph*{\glspl{ssm} and Mamba for \gls{eeg}}
\glspl{ssm} model sequence dynamics through latent state evolution governed by linear dynamical systems, enabling efficient long-range temporal modeling. Mamba~\cite{mamba} introduced input-dependent selection mechanisms that provide content-aware processing while maintaining linear-time complexity. For EEG signals, bidirectional variants are particularly important, as bidirectional token interactions enable richer contextual modeling of transient and non-stationary patterns~\cite{biomamba}. Bi-Mamba+~\cite{bimamba} combines forward and backward processing through gated fusion, and FEMBA~\cite{femba} demonstrated that \gls{bimamba} blocks can match Transformer performance on clinical \gls{eeg} benchmarks while achieving up to $3.5\times$ lower computational cost.

\paragraph*{Topology-Invariant Encoding}
\gls{eeg} datasets exhibit topological heterogeneity: electrode counts range from 20 (clinical bipolar) to 256 (research high-density), and montage conventions vary across institutions. Naive approaches either train separate models per configuration or retain shared electrodes, discarding substantial data. Several strategies address this challenge. LaBraM~\cite{labram} models joint spatio-temporal attention by flattening channel and patch dimensions, but incurs $\mathcal{O}((S \cdot C)^2)$ complexity. REVE~\cite{reve} instead uses 4D Fourier-based encodings to jointly represent electrode coordinates and temporal patches with complexity $\mathcal{O}(S \cdot C)$. LUNA~\cite{luna} similarly uses 3D spatial coordinate encodings, but further introduces learned-query cross-attention to project arbitrary electrode layouts into a fixed latent space with linear-in-channels complexity.

\paragraph*{\gls{ssl} Pre-training Objectives}
Current \gls{eeg} foundation models primarily rely on masked reconstruction, where random input patches are masked for the model to reconstruct them~\cite{femba,luna,labram}. Recently, \gls{lejepa}~\cite{balestriero2025lejepaprovablescalableselfsupervised} emerged as an alternative that promotes well-conditioned embedding distributions through (1) the \gls{sigreg}, which regularizes embeddings toward an isotropic Gaussian target using the Epps--Pulley statistical test~\cite{eepspulley} on random 1D projections, and (2) a \gls{jepa} predictive loss aligning local and global views of the input. Unlike prior approaches based on teacher--student schemes or exponential moving averages, \gls{lejepa} relies on a small set of hyperparameters, including the regularization scale $\lambda$ and the number of \gls{sigreg} projection slices. While effective on image and video data, its application to biosignal time series remains unexplored.

\textbf{LuMamba} combines LUNA's topology-invariant encoding with FEMBA's efficient \gls{bimamba} blocks, and provides the first adaptation of \gls{lejepa} to \gls{eeg}. We show that \gls{lejepa}'s distributional regularization, together with reconstruction-based pre-training, yield complementary benefits: reconstruction promotes structured latent spaces, while \gls{lejepa} improves cross-montage generalization.

\section{Methodology}
\label{sec:methods}
\begin{figure*}[t]
    \centering
    \includegraphics[width=1\textwidth]{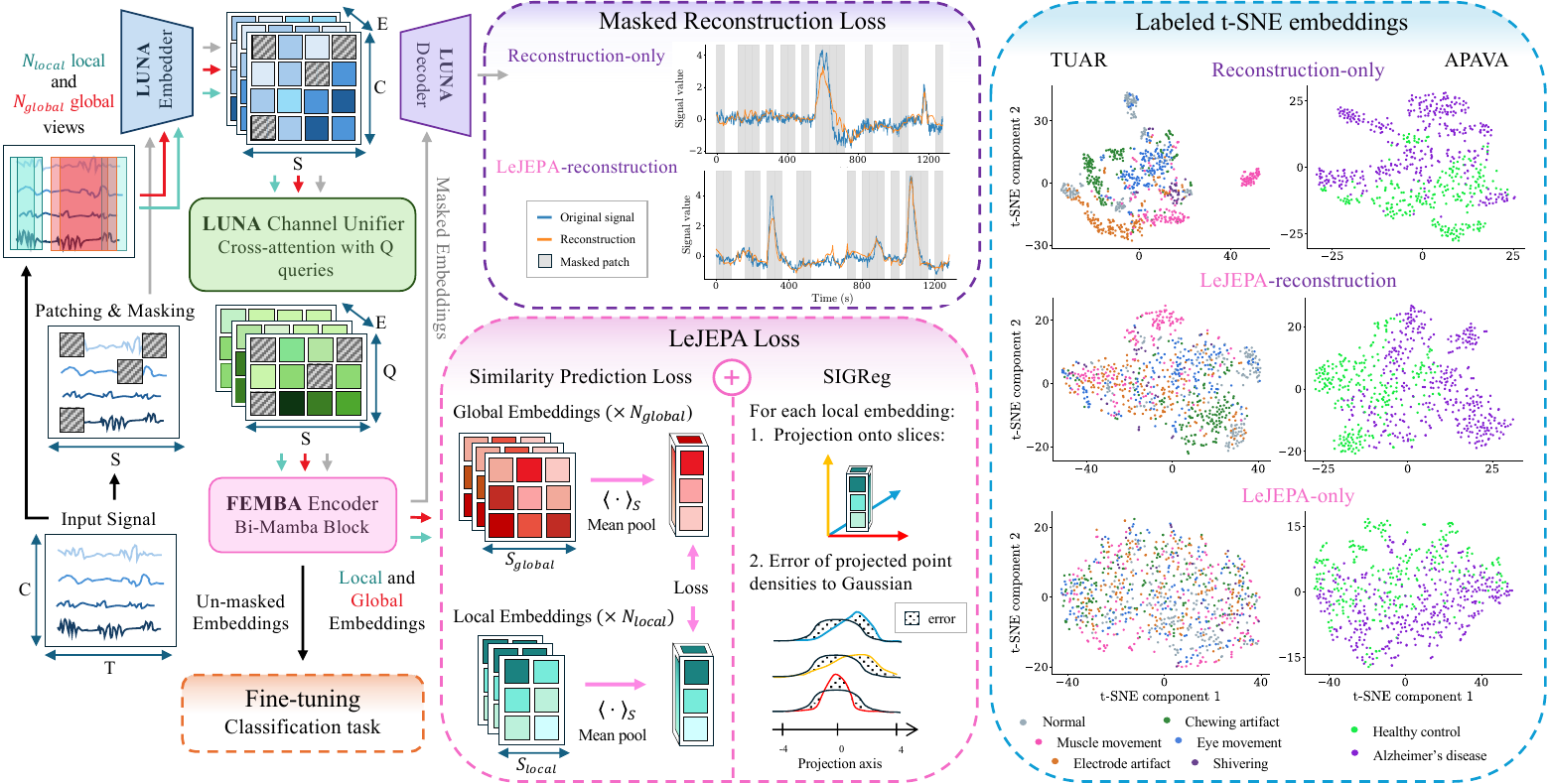}
    \captionsetup{font=footnotesize}
    \caption{Overview of the LuMamba architecture. \gls{tsne} of TUAR and APAVA embeddings compare reconstruction-only, \gls{lejepa}-reconstruction, and \gls{lejepa}-only strategies at the end of pre-training.}
    \label{fig:lumamba_architecture}
    \vspace{-0.5cm}
\end{figure*}
\subsection{Dataset}
We pretrain on the \gls{tueg} corpus, comprising approximately $21{,}600$ hours of recordings from more than $14{,}000$ patients, recorded using the $10$--$20$ system with $20$--$22$ channels~\cite{tueg}. Downstream experiments rely on task-specific subsets from the \gls{tuh} family. \textbf{\gls{tuab}} addresses normal/abnormal classification ($2{,}329$ subjects). \textbf{\gls{tuar}} targets multi-class artifact detection ($213$ subjects), following task definitions in~\cite{femba,luna}. \textbf{\gls{tusl}} ($38$ patients) covers the classification of slowing events, seizures, complex background, and normal \gls{eeg} activity~\cite{tueg}.

Motivated by recent work on biosignal \gls{ssm}s~\cite{biomamba}, we consider two disease-specific datasets: \textbf{APAVA}~\cite{apava} for \gls{ad} detection ($23$ patients; $16$ channels) and \textbf{TDBrain}~\cite{TDBrain} for \gls{pd} detection ($72$ patients; $26$ channels).

\subsection{Preprocessing}
We apply a preprocessing pipeline across all pre-training and fine-tuning datasets. Signals are bandpass-filtered between $0.1$ and $75$~Hz and notch-filtered at $50$~Hz, resampled to $256$~Hz, and segmented into non-overlapping $5$-s windows. For the TDBrain \gls{pd} subset, a $1.25$-s window is used to match the baseline preprocessing~\cite{biomamba}.
\subsection{Model Architecture}
Our architecture balances performance and efficiency with a compact footprint of $4.6$ million parameters, structured into four primary components as in Figure~\ref{fig:lumamba_architecture}.
\newcommand{\block}[1]{\textbf{#1:}\ }

\block{Encoder (LUNA~\cite{luna})} The input $x \in \mathbb{R}^{B \times C \times T}$ is tokenized into $S=T/P$ patches. Each patch is projected into $\mathbb{R}^E$ by fusing 1D-convolutional temporal features, \gls{fft}-based spectral features, and 3D electrode positional encodings, yielding a tensor $X_{tok} \in \mathbb{R}^{(B \cdot S) \times C \times E}$.

\block{Channel Unification (LUNA~\cite{luna})} We apply a cross-attention module where $Q$ learnable queries attend to the channel dimension $C$. This produces a montage-robust latent representation $X_{lat} \in \mathbb{R}^{(B \cdot S) \times Q \times E}$, decoupling the encoder's internal dimensionality from the input electrode configuration.

\block{Temporal Modeling (FEMBA~\cite{femba})}
$X_{lat}$ is reshaped to $B \times S \times (Q \cdot E)$ and processed by two \gls{bimamba} blocks. These blocks leverage \gls{ssm}s to capture bidirectional dependencies.

\block{Decoder and Classifier (LUNA~\cite{luna})}
For pre-training, a cross-attention decoder maps $X_{lat}$ back to the original channel space using $C$ learnable queries. For fine-tuning, the decoder is replaced by a  classification head.

\subsection{Self-supervised Pre-training}

\begin{table*}[t]
\centering
\captionsetup{font=footnotesize}
\caption{LuMamba pre-training strategy comparison on \gls{tuh} tasks and mental condition tasks TDBrain and APAVA, evaluating \gls{balacc}, \gls{auroc}, and \gls{aupr}.}
\label{tab:results_pretraining_strategy}
\setlength{\tabcolsep}{4pt}
\resizebox{\textwidth}{!}{
\begin{tabular}{@{}cccccccccccc@{}}
\toprule
\textbf{Pre-training strategy} &
\multicolumn{3}{c}{\textbf{\gls{tuab}}} &
\multicolumn{2}{c}{\textbf{\gls{tuar}}} &
\multicolumn{2}{c}{\textbf{\gls{tusl}}} &
\multicolumn{2}{c}{\textbf{TDBrain}} &
\multicolumn{2}{c}{\textbf{APAVA}} \\
\cmidrule(lr){2-4}
\cmidrule(lr){5-6}
\cmidrule(lr){7-8}
\cmidrule(lr){9-10}
\cmidrule(lr){11-12}
 &
\gls{balacc} (\%) & \gls{auroc} & \gls{aupr} &
\gls{auroc} & \gls{aupr} &
\gls{auroc} & \gls{aupr} &
\gls{auroc} & \gls{aupr} &
\gls{auroc} & \gls{aupr} \\
\midrule
Reconstruction-only &
80.36 $\pm$ 00.39 & 0.8782 $\pm$ 0.0050 & 0.8719 $\pm$ 0.0067 &
\textbf{0.914 $\pm$ 0.007} & \textbf{0.510 $\pm$ 0.020} &
\textbf{0.708 $\pm$ 0.036} & \textbf{0.289 $\pm$ 0.013} &
0.961 $\pm$ 0.003 & 0.958 $\pm$ 0.004 &
0.714 $\pm$ 0.261 & 0.765 $\pm$ 0.209\\
\gls{lejepa}-only &
80.02 $\pm$ 0.50 & 0.8747 $\pm$ 0.0038  & 0.8817 $\pm$  0.0038&
0.891 $\pm$ 0.006 & 0.502 $\pm$ 0.009 &
0.540 $\pm$ 0.038 & 0.261 $\pm$ 0.014 &
0.930 $\pm$ 0.033 & 0.938 $\pm$ 0.015 &
0.816 $\pm$ 0.102 & 0.819 $\pm$ 0.109 \\
\gls{lejepa}-reconstruction &
\textbf{80.99 $\pm$ 0.22} & \textbf{0.8825 $\pm$ 0.0038} & \textbf{0.8918 $\pm$ 0.0032} &
0.896 $\pm$ 0.020 & 0.490 $\pm$ 0.023 &
0.660 $\pm$ 0.053 & 0.272 $\pm$ 0.011 &
\textbf{0.961 $\pm$ 0.006 }& \textbf{0.960 $\pm$ 0.007} &
\textbf{0.955 $\pm$ 0.018} & \textbf{0.970 $\pm$ 0.012} \\
\bottomrule
\end{tabular}
}
\vspace{-0.3cm}
\end{table*}
We investigate whether self-supervised learning for \gls{eeg} can go beyond the masked reconstruction paradigm. To this end, we adapt \gls{lejepa}~\cite{balestriero2025lejepaprovablescalableselfsupervised}, originally proposed for images, to \gls{eeg} signals. By encouraging more isotropic embedding distributions, \gls{lejepa} can reduce sensitivity to classifier variability and lower downstream prediction risk~\cite{balestriero2025lejepaprovablescalableselfsupervised}. We evaluate whether this objective yields more informative representations than reconstruction alone by pre-training LuMamba with masked reconstruction, \gls{lejepa}, and their combination.
\subsubsection{Masked reconstruction}
Following~\cite{femba,luna}, a 60\% subset of the input patches is randomly masked. The encoder is trained to extract meaningful \gls{eeg} representations that enable reconstruction of the missing patches from the visible context.
\subsubsection{\gls{lejepa}}
\gls{lejepa} exploits two components:
\begin{itemize}
    \item \textbf{\gls{sigreg}}, which regularizes latent embeddings toward an isotropic Gaussian distribution.
    \item A \textbf{\gls{jepa} predictive loss}, which aligns local views of an input with global views containing a wider input context.
\end{itemize}
Inspired by the original \gls{lejepa} view construction from cropped image patches, we adapt this idea to the temporal domain by forming local and global views through random temporal windows extracted from the input signal.

Given $x \in \mathbb{R}^{B \times C \times T}$, we sample $N_{global}$ global and $N_{local}$ local temporal windows of sizes $T_{global}$ and $T_{local}$, where $T_{global} > T_{local}$.
In our setup, $N_{global}=2$ and $N_{local}=4$.

We embed $x_{local}$ and $x_{global}$ with the LuMamba encoder, obtaining, after sequence flattening, embeddings of shapes $v_{local} \in \mathbb{R}^{N_{local} \times B \times (Q \cdot E)}$ and $v_{global} \in \mathbb{R}^{N_{global} \times B \times (Q \cdot E)}$.

The \textbf{\gls{jepa} predictive loss} is then given by:
\[
\frac{1}{N_{\text{local}}} \sum_{i=1}^{N_{\text{local}}}
\lVert \boldsymbol{\mu}_{\text{global}} - v_{\text{local},i} \rVert_2^2,
\quad
\boldsymbol{\mu}_{\text{global}}
= \frac{1}{N_{\text{global}}} \sum_{j=1}^{N_{\text{global}}} v_{\text{global},j}.
\]

Finally, \textbf{\gls{sigreg}} projects the $v_{global}$ and $v_{local}$ embeddings into a lower-dimensional space defined by $M=300$ projection slices. For each dimension, the Epps--Pulley test~\cite{eepspulley} measures the discrepancy between the \gls{ecf} of the embeddings and the \gls{cf} of an isotropic Gaussian target distribution.

\subsection{Fine-Tuning}
\subsubsection{Classifier Architectures}

After pre-training, the decoder is replaced with a lightweight Mamba-based classification head from FEMBA~\cite{femba}, consisting of a non-bidirectional Mamba-enhanced linear layer with $536$K parameters. This choice favors computational efficiency over the transformer-based cross-attention classifier used in LUNA~\cite{luna}.

\subsubsection{Downstream tasks}
LuMamba is evaluated on five \gls{eeg} datasets: three \gls{tuh} benchmarks (\gls{tuab}, \gls{tusl}, \gls{tuar}) and two datasets with unseen channel montages (TDBrain and APAVA). All five datasets are strictly excluded from the pre-training stage. Following standard evaluation protocols~\cite{femba,luna}, \gls{tuab} uses the official split, while \gls{tusl} and \gls{tuar} use an 80/10/10 train/val/test partition. For APAVA and TDBrain, we adopt subject-disjoint splits from the literature~\cite{wang2024medformer,biomamba}: 15/4/4 and 34/8/8 subjects for train/val/test, respectively. During fine-tuning, the entire architecture remains unfrozen. Training is limited to 30 epochs, and results are averaged over three runs with fixed random seeds.

\section{Results}
\label{sec:results}

\subsection{Assessing benefits of \gls{lejepa}}
\subsubsection{\gls{tsne}}
We analyze the impact of \gls{lejepa} on the latent space using \gls{tsne} projections\cite{tSNE}. Despite the absence of class supervision, reconstruction-only pre-training yields increasingly well-separated clusters on TUAR and APAVA (Figure~\ref{fig:lumamba_architecture}). In contrast, combining \gls{lejepa} with reconstruction preserves signal reconstruction quality but results in more diffuse and isotropic embeddings (Figure~\ref{fig:lumamba_architecture}).
As the contribution of \gls{lejepa} in pre-training increases, apparent cluster separation consistently decreases. Rather than promoting visually structured latent spaces, \gls{lejepa} smooths the representation geometry, shifting toward more uniform representations.
\subsubsection{Query attention maps}
\begin{figure}[h]
    \vspace{-0.3cm}
    \centering
    \includegraphics[width=.4\textwidth]{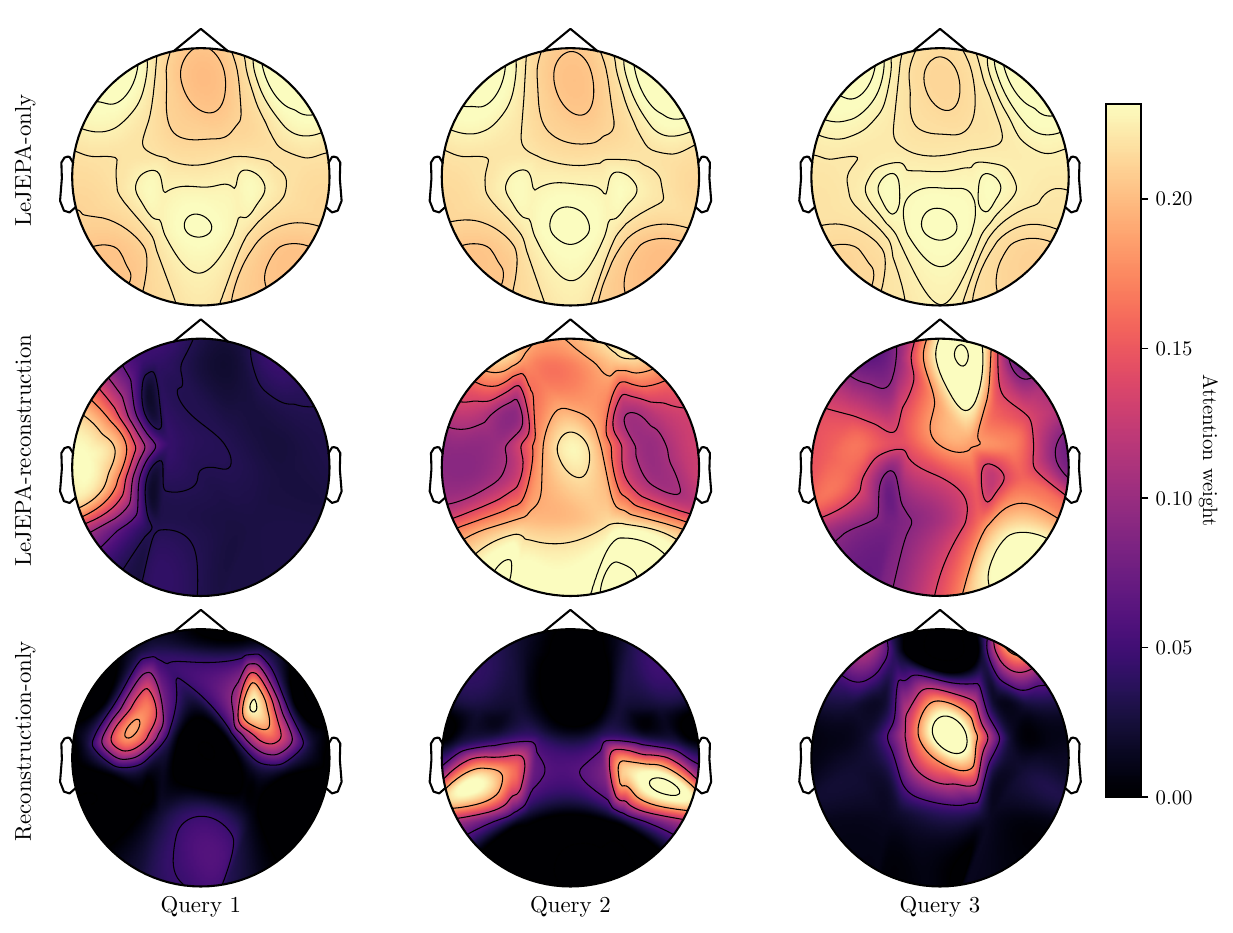}
    \captionsetup{font=footnotesize}
    \caption{Query attention patterns on scalp topology, comparing \gls{lejepa}, \gls{lejepa}-reconstruction, and reconstruction-only.}
    \label{fig:pretrain_comparison}
    \vspace{0.1cm}
\end{figure}

To further examine the role of isotropy in the embedding space, we visualize channel-unification attention patterns projected onto a head layout (Figure~\ref{fig:pretrain_comparison}). \gls{lejepa} alone produces homogeneous queries with limited spatial differentiation, whereas reconstruction-only yields sharp attention peaks surrounded by low-attention regions, indicating highly localized topology modeling.
\gls{lejepa}--reconstruction exhibits an intermediate behavior, with smoother transitions between attention responses while preserving spatial contrast. This graded attention distribution may better capture interactions across brain regions associated with higher-order cognitive patterns~\cite{connectivityanalysis, EEGbrainfunctionalconnectivity}. Moderate isotropy therefore appears to relax reconstruction-induced locality, enabling more robust representations under sparse spatial sampling and potentially improving transferability.
\subsubsection{Downstream tasks}
To contextualize these qualitative observations, we compare the three pre-training strategies on downstream performance. The more structured representations of reconstruction-only pre-training translate into slightly higher performance on in-distribution \gls{tuh} tasks, outperforming \gls{lejepa}--reconstruction by $1$--$2\%$ on TUAR and up to $4\%$ on TUSL.
However, this increased clustering appears to reduce transferability across heterogeneous channel topologies (Table ~\ref{tab:results_pretraining_strategy}). In contrast, the mixed \gls{lejepa}--reconstruction objective shows improved generalization on three out of five tasks, including TDBrain and APAVA, both involving electrode montages unseen during pre-training. This effect is particularly visible on the sparse 16-channel APAVA topology, where incorporating \gls{lejepa} improves \gls{ad} detection by over \textbf{$20\%$} compared to reconstruction-only.
Overall, while reconstruction promotes visually clustered representations that favor in-distribution performance, \gls{lejepa} regularizes the representation space, improving robustness and cross-task transferability. The two objectives therefore play complementary roles when paired, providing a reliable trade-off between representation structure and generalization. The mixed \gls{lejepa}--reconstruction objective is subsequently used for benchmarking.
\begin{table}[b!]
    \vspace{-0.4cm}
    \centering
    \captionsetup{font=footnotesize}
    \caption{Performance Comparison on TUAB}
    \label{tab:results_tuab_soa}
    \resizebox{\columnwidth}{!} {
    \setlength{\tabcolsep}{6pt} 
    \begin{tabular}{@{}lcccc@{}}
        \toprule
        \textbf{Model} & \textbf{Size} & \textbf{\gls{balacc} (\%)} & \textbf{\gls{aupr}} & \textbf{\gls{auroc}}\\
    \midrule
    BENDR \cite{bendr} & 0.39M & 76.96 $\pm$ 3.98 & - & 0.8397 $\pm$ 0.0344 \\
    EEGFormer-Base \cite{eegformer_foundation} & 2.3M & - & 0.8670 $\pm$ 0.0020 & 0.8670 $\pm$ 0.0030 \\
    BIOT \cite{biot} & 3.2M & 79.59 $\pm$ 0.57 & 0.8692 $\pm$ 0.0023 & 0.8815 $\pm$ 0.0043 \\
    LaBraM-Base \cite{labram} & 5.9M & \textbf{81.40 $\pm$ 0.19} & \textbf{0.8965 $\pm$ 0.0016} & \textbf{0.9022 $\pm$ 0.0009} \\    
    LUNA-Base \cite{luna} & 7M  & 80.63 $\pm$ 0.08 & 0.8953 $\pm$ 0.0016  & 0.8868 $\pm$ 0.0015 \\
    CEReBro \cite{cerebro} & 3.6M  & 79.40 $\pm$ 0.19 & 0.8763 $\pm$ 0.0031  & 0.8749 $\pm$ 0.0033 \\
    \midrule
    \textbf{LuMamba-Tiny} & 4.6M  &  80.99 $\pm$ 0.22 & 0.8918 $\pm$ 0.0032 & 0.8825 $\pm$ 0.0038 \\
    \bottomrule
   \end{tabular}}
   \vspace{0.4cm}
\end{table}
\subsection{Comparison to \gls{soa} baselines}
\begin{table}[b!]
\vspace{-0.4cm}
\centering
\captionsetup{font=footnotesize}
\caption{Performance comparison on TDBrain and APAVA}
\label{tab:new_montages}
\resizebox{\columnwidth}{!} {
\setlength{\tabcolsep}{2pt} 
\begin{tabular}{@{}lccccccc@{}}
\toprule
\textbf{Model} & \textbf{Size} & \multicolumn{2}{c}{\textbf{TDBrain}} & \multicolumn{2}{c}{\textbf{APAVA}} \\
\cmidrule(lr){3-4} \cmidrule(lr){5-6}
 &  & \gls{auroc} & \gls{aupr} & \gls{auroc} & \gls{aupr} \\
\midrule
\textbf{Supervised Models} \\
Medformer \cite{wang2024medformer} & 8M & 0.959 $\pm$ 0.007 & 0.960 $\pm$ 0.006 & 0.811 $\pm$ 0.046 & 0.816 $\pm$ 0.429 \\
BioMamba \cite{biomamba} & 1M & \textbf{0.994 $\pm$ 0.005} & \textbf{0.994 $\pm$ 0.005} & 0.938 $\pm$ 0.014 & 0.935 $\pm$ 0.014 \\
\midrule
\textbf{LuMamba-Tiny} & 4.6M & 0.961 $\pm$ 0.006 & 0.960 $\pm$ 0.007 & \textbf{0.955 $\pm$ 0.018} & \textbf{0.970 $\pm$ 0.012} \\

\bottomrule
\end{tabular}
}
\vspace{-0.4cm}
\end{table}
\begin{table}[t!]
\centering
\captionsetup{font=footnotesize}
\caption{Performance comparison on TUAR and TUSL}
\label{tab:results_tusl_tuar}
\resizebox{\columnwidth}{!} {
\setlength{\tabcolsep}{2pt} 
\begin{tabular}{@{}lccccccc@{}}
\toprule
\textbf{Model} & \textbf{Size} & \multicolumn{2}{c}{\textbf{TUAR}} & \multicolumn{2}{c}{\textbf{TUSL}} \\
\cmidrule(lr){3-4} \cmidrule(lr){5-6}
 & & \gls{auroc} & \gls{aupr} & \gls{auroc} & \gls{aupr}  \\
\midrule
EEGFormer-Base~\cite{eegformer_foundation} & 2.3M & $0.847 \pm 0.014$ & $0.483 \pm 0.026$ & $0.713 \pm 0.010$ & \textbf{0.393 $\pm$ 0.003} \\
FEMBA-Tiny~\cite{femba} & 7.8M  & \textbf{0.918 $\pm$ 0.003} & \textbf{0.518 $\pm$ 0.002} & 0.708 $\pm$ 0.005 & 0.277 $\pm$ 0.007 \\
LUNA-Base ~\cite{luna}  & 7M & $0.902 \pm 0.011$ & $0.495 \pm 0.010$ & \textbf{0.767 $\pm$ 0.023} & $0.301 \pm 0.003$ \\
\midrule
\textbf{LuMamba-Tiny}  & 4.6M & 0.896 $\pm$ 0.020 & 0.490 $\pm$ 0.023 & 0.660 $\pm$ 0.053 & 0.272 $\pm$ 0.011 \\
\bottomrule
\end{tabular}
}
\vspace{-0.4cm}
\end{table}

On TUAB, LuMamba achieves performance similar to LUNA and approaches LaBraM in \gls{balacc} and \gls{auroc},  with overlapping confidence intervals. However, LuMamba trails LaBraM  in \gls{aupr} by approximately $-$2\%. Overall, among self-supervised models using a comparable evaluation setup, LuMamba shows competitive performance, with the corresponding results summarized in Table \ref{tab:results_tuab_soa}.
On mental condition tasks with unseen montages, LuMamba demonstrates strong generalization. On APAVA (\gls{ad} detection), it achieves approximately $+$4\% \gls{aupr} over the previous \gls{soa}, while on TDBrain (\gls{pd} detection) performance remains comparable to Medformer and slightly below BioMamba (Table \ref{tab:new_montages}).
On TUAR and TUSL, LuMamba underperforms task-specific \gls{soa} methods (Table \ref{tab:results_tusl_tuar}), with a more pronounced gap on TUSL, which is known to be highly class-imbalanced~\cite{tueg}. We attribute this behavior not to architectural limitations, but to our methodological focus on generalization and transferable pre-training. Notably, results in Table~\ref{tab:results_pretraining_strategy} indicate that the TUSL performance gap is not due to the architectural design, as the same architecture with reconstruction-based pretraining achieves near-\gls{soa} performance; rather, it stems from our choice of a generalization-oriented pretraining strategy, which is less favorable on highly imbalanced datasets such as TUSL.

\subsection{Computational efficiency and scalability}
We analyze the computational efficiency of LuMamba in terms of \gls{flops} with respect to sequence length (Figure~\ref{fig:flops}). Due to the linear-time complexity of the Mamba block, LuMamba consistently requires fewer \gls{flops} than attention-based foundation models at equal sequence lengths. At the maximum sequence length supported by each baseline before \gls{oom}, LuMamba requires \textbf{26.5$\times$} fewer \gls{flops} than LUNA, \textbf{377$\times$} fewer than LaBraM, and \textbf{3718$\times$} fewer than EEGFormer when evaluated at the same length, corresponding to absolute savings of up to several thousand G\gls{flops}.
In addition to length-wise efficiency, LuMamba exhibits improved scalability, supporting sequences that are \textbf{12.6$\times$} longer than LUNA, \textbf{501$\times$} longer than LaBraM, and \textbf{2.5$\times$} longer than EEGFormer before reaching the \gls{oom} limit.
These results indicate that replacing the transformer-based encoder in LUNA with a \gls{bimamba} encoder substantially improves both computational efficiency and sequence-length scalability while preserving the overall modeling framework.

\begin{figure}[h]
    \vspace{-0.4cm}
    \centering
    \begin{tikzpicture}
\def\oomx{25119}
\def\oomxlabram{631}
\def\oomxbiot{63096}
\def\oomxeegformer{125893}
\begin{axis}[
    width=0.99\columnwidth,
    height=0.55\columnwidth,
    xmode=log,
    ymode=log,
    log basis x=10,
    log basis y=10,
    xlabel={Sequence Length},
    ylabel={FLOPs (GFLOPs)},
    grid=major,
    mark size=1.0pt,
    minor tick length=0pt,
    xmin=0.7,
    xmax=4e5,   
    ymin=1e-3,
    ymax=1e7,
    ytick align=inside,
    every axis plot/.append style={semithick},
    legend pos=north west,
    tick label style={font=\tiny},
    label style={font=\small},
    legend style={font=\tiny},
]

\addplot[
    darkgray,
    mark=*,
] table[
    col sep=comma,
    x=patches_per_channel,
    y=gflops
] {Figures/luna_overleaf.csv};
\addlegendentry{LUNA}

\addplot[
    blue,
    mark=square*,
] table[
    col sep=comma,
    x=patches_per_channel,
    y=gflops
] {Figures/lumamba_overleaf.csv};
\addlegendentry{LuMamba}

\addplot[
    gray,
    mark=triangle,
] table[
    col sep=comma,
    x=patches_per_channel,
    y=gflops
] {Figures/labram_overleaf.csv};
\addlegendentry{LaBraM}

\addplot[
    Periwinkle,
    mark=triangle*,
] table[
    col sep=comma,
    x=patches_per_channel,
    y=gflops
] {Figures/eegformer_overleaf.csv};
\addlegendentry{EEGFormer}

\addplot[
    red,
    dashed,
] coordinates {(\oomx,1e-3) (\oomx,1e7)};

\addplot[
    red,
    dashed,
] coordinates {(\oomxlabram,1e-3) (\oomxlabram,1e7)};

\addplot[
    red,
    dashed,
] coordinates {(\oomxeegformer,1e-3) (\oomxeegformer,1e7)};
\node[red, anchor=north east] at (axis description cs:0.88,0.3) {Out Of Memory};
\end{axis}
\end{tikzpicture}
    \captionsetup{font=footnotesize}
     \caption{FLOPs vs sequence length for different \gls{eeg} foundation models. Dashed lines denote the \gls{oom} limit on an NVIDIA A100-64GB GPU (batch size = 1).}
    \label{fig:flops}
\vspace{-0.5cm}
\end{figure}

\section{Conclusion}\label{ch:conclusion}

We introduced LuMamba, an \gls{eeg} foundation model that extends \gls{ssm}-based learning to heterogeneous channel topologies by combining LUNA's topology-invariant encoding with FEMBA's efficient \gls{bimamba} blocks.
With only 4.6M parameters, LuMamba achieves linear sequence-length scaling, requiring 377$\times$ fewer \gls{flops} than LaBraM and supporting 12$\times$ longer sequences before reaching memory limits.
We further introduced a mixed \gls{lejepa}--reconstruction pre-training objective that regularizes the latent space toward isotropic distributions, improving robustness to distribution shifts.
LuMamba achieves performance comparable to established \gls{tuh} baselines while exhibiting stronger generalization on unseen montages, most notably a $+$4\% \gls{aupr} improvement on \gls{ad} detection.
Future work will expand downstream evaluations and scale the pre-training corpus to further assess the generalizability of the proposed framework.

\section*{Acknowledgment}
 We acknowledge ISCRA for awarding this project access to the LEONARDO supercomputer, owned by the EuroHPC Joint Undertaking, hosted by CINECA (Italy). This work was supported by a grant from the Swiss National Supercomputing Centre (CSCS) under project ID lp12 and lp160 on Alps.

\bibliographystyle{IEEEtran}
\bibliography{bib}

\end{document}